\documentclass[a4paper,conference]{IEEEtran}
\IEEEoverridecommandlockouts
\ifCLASSINFOpdf
\usepackage{times}
\usepackage{helvet}
\usepackage{courier}

\usepackage{amsmath,amssymb} 
\usepackage{color}
\usepackage{multicol}
\usepackage{multirow}
\usepackage{array}
\usepackage{amssymb}
\usepackage{subfigure}
\usepackage{comment}
\usepackage{booktabs}
\usepackage{graphicx}
\usepackage{makecell}

\let\OLDthebibliography\thebibliography
\renewcommand\thebibliography[1]{
	\OLDthebibliography{#1}
	\setlength{\parskip}{0pt}
	\setlength{\itemsep}{0pt plus 0.3ex}
}

\DeclareRobustCommand*{\IEEEauthorrefmark}[1]{%
	\raisebox{0pt}[0pt][0pt]{\textsuperscript{\footnotesize\ensuremath{#1}}}}

\begin{document}\sloppy
	\def\x{{\mathbf x}}
	\def\L{{\cal L}}

	\title{Efficient Human Pose Estimation by Learning Deeply Aggregated Representations}
	\author{
		\IEEEauthorblockN{
			Zhengxiong Luo\IEEEauthorrefmark{1}\IEEEauthorrefmark{2}\IEEEauthorrefmark{4}\IEEEauthorrefmark{5}\IEEEauthorrefmark{6},
			Zhicheng Wang\IEEEauthorrefmark{1},
			Yuanhao Cai\IEEEauthorrefmark{1}\IEEEauthorrefmark{3},
			Guanan Wang\IEEEauthorrefmark{1}\IEEEauthorrefmark{6}\\
			Yan Huang\IEEEauthorrefmark{4}\IEEEauthorrefmark{5}\IEEEauthorrefmark{6}, 
			Liang Wang\IEEEauthorrefmark{4}\IEEEauthorrefmark{5}\IEEEauthorrefmark{6}, 
			Erjin Zhou\IEEEauthorrefmark{1},
			Tieniu Tan\IEEEauthorrefmark{4}\IEEEauthorrefmark{5},
			Jian Sun\IEEEauthorrefmark{1}\\
		}
		\IEEEauthorblockA{
			\IEEEauthorrefmark{1} Megvii Inc \quad
			\IEEEauthorrefmark{2} University of Chinese Academy of Sciences (UCAS) \quad
			\IEEEauthorrefmark{3} Tsinghua Shenzhen International Graduate School\\
			\IEEEauthorrefmark{4} Center for Research on Intelligent Perception and Computing (CRIPAC)\quad
			\IEEEauthorrefmark{5} National Laboratory of Pattern Recognition (NLPR)\\
			\IEEEauthorrefmark{6} Institute of Automation, Chinese Academy of Sciences (CASIA)\\
			\{luozhengxiong, wangzhicheng, caiyuanhao,wangguanan,zej,sunjian\}@megvii.com \\ 
			\{yhuang, wangliang, tnt\}@nlpr.ia.ac.cn
		}
	}
	
	\maketitle
	
	\begin{abstract}
		
		In this paper, we propose an efficient human pose estimation network (DANet) by learning deeply aggregated representations. Most existing models explore multi-scale information mainly from features with different spatial sizes. Powerful multi-scale representations usually rely on the cascaded pyramid framework. This framework largely boosts the performance but in the meanwhile makes networks very deep and complex. Instead, we focus on exploiting multi-scale information from layers with different receptive-field sizes and then making full of use this information by improving the fusion method. Specifically, we propose an orthogonal attention block (OAB) and a second-order fusion unit (SFU). The OAB learns multi-scale information from different layers and enhances them by encouraging them to be diverse. The SFU adaptively selects and fuses diverse multi-scale information and suppress the redundant ones. This could maximize the effective information in final fused representations. With the help of OAB and SFU, our single pyramid network may be able to generate deeply aggregated representations that contain even richer multi-scale information and have a larger representing capacity than that of cascaded networks. Thus, our networks could achieve comparable or even better accuracy with much smaller model complexity. Specifically, our \mbox{DANet-72} achieves $70.5$ in AP score on COCO test-dev set with only $1.0G$ FLOPs. Its speed on a CPU platform achieves $58$ Persons-Per-Second~(PPS). The source code will be publicly available for further research.
		
	\end{abstract}

	\section{Introduction}
	
\begin{figure*}[t]
		\centering
		\includegraphics[width=\linewidth]{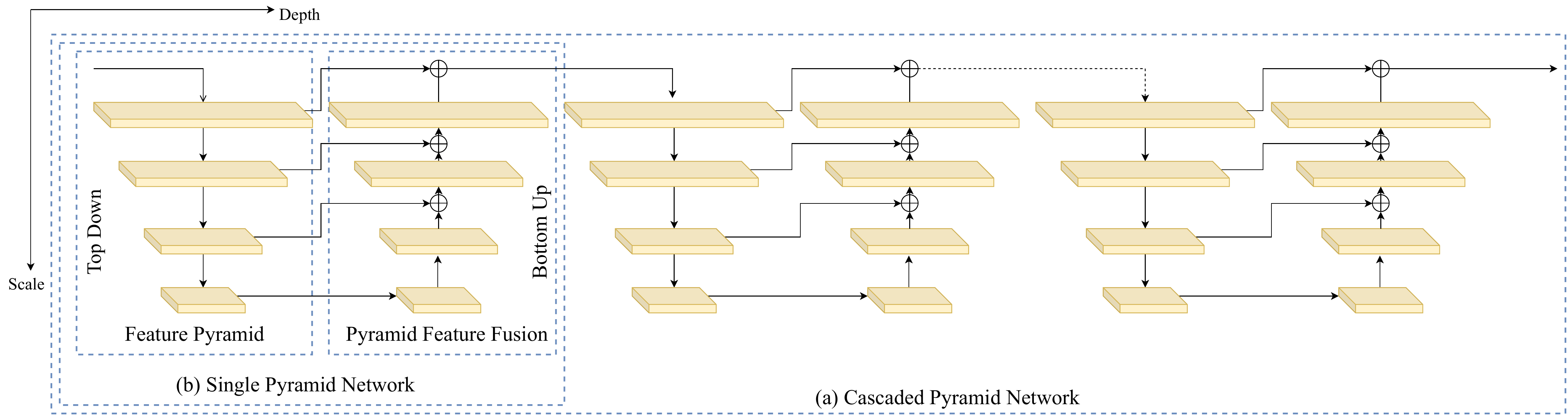}
		\vspace{-0.03\linewidth}
		\caption{Different framework for human pose pose estimation. (a) Cascaded Pyramid Network, (b) Single Pyramid Network.} 
		\label{cascade_u}
		\vspace{-0.03\linewidth}
	\end{figure*}

	Human pose estimation (HPE) aims to localize all skeletal keypoints of given persons in a single RGB image. It has wide applications in human activity recognition, human-computer interaction, animation, etc. To attain sub-pixel accuracy, HPE needs both local spatially precise and global semantically discriminant representations. Thus rich multi-scale information is quite important for HPE. Towards this motivation, as shown in Figure~\ref{cascade_u} (a), most existing models~\cite{stacked,mspn,yang2017learning,fast_human,rsn} use a cascaded pyramid framework to repeatedly extract and fuse pyramid features, i.e. features with different spatial sizes. This framework largely boosts the performance of HPE but makes these networks very deep and complex.
	
	As shown in Figure~\ref{cascade_u}~(b), a single pyramid network is much simpler. However, without repeatedly extracting and fusing pyramid features, it can only generate limited multi-scale information from features of four different spatial sizes, which affects the accuracy a lot. This motivates us to think: if features of every single spatial size could independently learn rich multi-scale information, and further this information can be fused more effectively, even a single pyramid network may be able to achieve remarkable accuracy. In this paper, we propose a novel network, namely deeply aggregated network (DANet). It aggregates layers with different receptive-fields sizes, in which way multi-scale information could be extracted within features of the same spatial size. And the extracted information is further fused by a second-order fusion, which can maximize the effective representations in the final fused features.
	
	\begin{figure}[t]
		\centering
		\includegraphics[width=\linewidth]{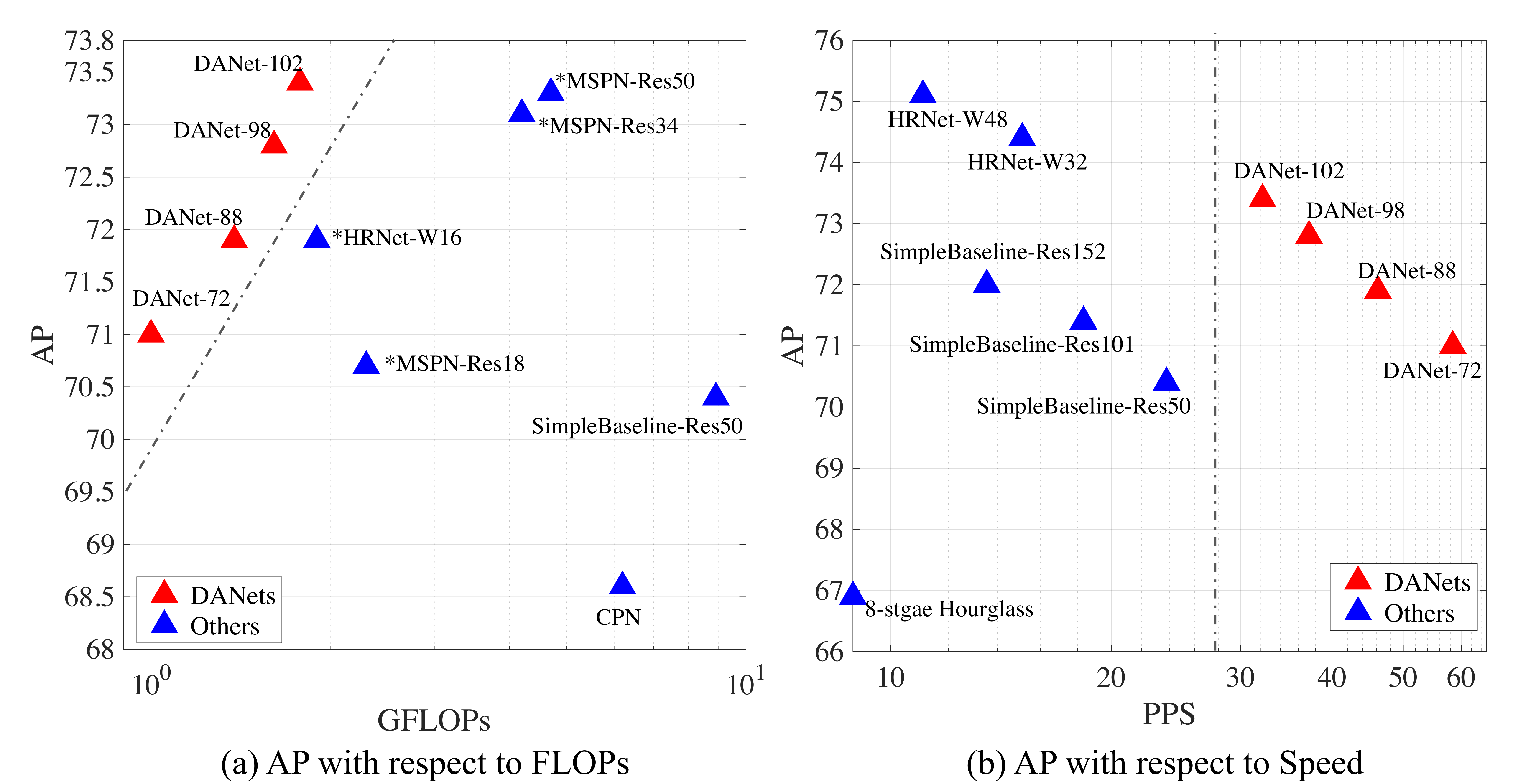}
		\caption{AP score of different models on COCO validation set with respect to (a) flops and (b) speed.}
		\label{speed}
	\end{figure} 
	
	Specifically, DANet has two key components, which are orthogonal attention block (OAB) and second-order fusion unit (SFU). The OAB aggregates different layers via a densely connected architecture similar to~\cite{densenet} and further enhances the multi-scale information from two views. A mask attention unit is designed to explicitly encourage different layers to be diverse by masking the representations of input features out from output features. Thus, the top features include less repeated information from the bottom layers. A channel attention unit is used to improve the nonlinearity of each layer and thus help the output features to be more different from the input ones. The SFU adaptively selects and fuses multi-scale information from multifarious features. It highlights the diversities of features and suppresses their similarities via adaptive weights, which are obtained by modeling dependencies among different features. As a result, the effective information in the final fused features can be maximized. With the help of OAB and SFU, our single pyramid network may be able to generate deeply aggregated representations, which even contain more powerful multi-scale information and have a larger representing capacity than that of cascaded networks. Consequently, comparable or even better accuracy can be achieved with much smaller model complexity.
	
	Our contributions can be summarized as four points:
	
	\begin{itemize}
		\item[1.] Towards efficient human pose estimation (HPE), we propose a deeply aggregated network (DANet), which focuses on exploring the multi-scale information within features of the same spatial sizes and improving the method that is used to fuse multi-scale information. As a result, our single pyramid network can generate deeply aggregated representations which can produce better accuracy than cascaded networks but with much smaller model complexity.
		
		\item[2.] We propose a novel orthogonal attention block (OAB), which explores and enhances multi-scale information from layers with different receptive-field sizes. It encourages different layers to be more diverse and enables layer-wise aggregated features to have a higher capacity. 
		
		\item [3.] We propose a novel second-order fusion unit (SFU), which adaptively selects and fuses multi-scale information from multifarious features by modeling their mutual dependencies. In this way, it maximizes the effective information in the final fused features.
		
		\item[4.] Experiments on both COCO and MPII datasets show that DANets could achieve a better balance between cost and effectiveness than existing models. Specifically, our DANet-72 can achieve $70.5$ in AP score on COCO test-dev set with only $1.0G$ FLOPs. Its speed on the CPU platform achieves about $58$ Persons-Per-Second~(PPS).
		
	\end{itemize}

	\section{Related Work}
	
	\subsection{Pyramid Fusion}
	
	Pyramid fusion is popular among models for human pose estimation (HPE), semantic segmentation~\cite{panet} and object detection \cite{fpn,dfpn,ssd,nasfpn,kong2018deep}. These tasks all require representations rich in multi-scale information. For human pose estimation and semantic segmentation, global information is needed for semantic discrimination and local information is used to determine spatial positions. While in object detection, scale-invariant representations are critical for hunting objects of various sizes. There are usually two paths in these models, i.e. the top-down path and the bottom-up path. In the top-down path,  a backbone such as ResNet~\cite{resnet} is usually used to extract the pyramid features. The feature sizes go from $1/4$ to $1/32$ of original images. 
	
	And in the bottom-up path, features with different spatial sizes are usually directly summed up to form the final representations. But simple summation ignores the dependencies between these features and thus limits the representing capacity of the final representations. In this paper, we improve this method by proposing a second-order fusion unit~(SFU), which adaptively selects and fuse multi-scale information from multifarious features by modeling their mutual dependencies. In this way, it maximizes the effective information in the final fused features.
	
	\subsection{Cascaded Pyramid Framework}
	
	\label{cascaded_pyramid_framework}
	For segmentation and detection, a single feature pyramid can usually satisfy the requirements of both tasks, even though the multi-scale information in a single pyramid is relatively limited because both tasks are indeed not so sensitive to sub-pixel accuracy as HPE. Sub-pixel localization needs much finer and richer multi-scale information. Consequently, HPE networks usually use a cascaded pyramid framework, which repeatedly extracts and fuse the feature pyramid and generates representations that can detect the sub-pixel difference. Since the proposal of Stacked Hourglass~\cite{stacked}, the cascaded pyramid framework has nearly become the dominant method in HPE~\cite{mspn,fast_human,scarb,ke2018multi}. Recently, HRNet proposed in~\cite{hrnet} uses a slightly different architecture that can maintain the high-resolution representations during the whole network. But in fact, it also repeatedly exchanges information between different scales. Single pyramid networks such as SimpleBaseline~\cite{cpm,cpn,simplebaseline} can only produce inferior results.
	
	The cascaded pyramid framework largely boosts the performance of HPE networks, but it makes these networks very deep and complex in the meanwhile. To produce relatively accurate results with a single pyramid network, we pay our attention to layer-wise aggregation and a more effective fusion method. With the help of diverse layer-wise aggregation and higher-order fusion, the proposed DANet can generate deeply aggregated representations that contain even richer multi-scale information and have a larger representing capacity than cascaded networks.
	
	\subsection{Densely Connected Architecture}
	\label{densely_connected_arch}
	
	Densely connected architecture is firstly proposed in~\cite{densenet} for image classification. It constantly concatenates newly generated features to previous ones. This inner structure makes it naturally have the property of multi-scale fusion~\cite{huang2017multi}. Compared with the fusion between features with different spatial sizes, this layer-wise fusion has much finer granularity. It also largely saves the computation because of feature reuse~\cite{densenet}. A similar idea can also be discovered in~\cite{shufflev2}, which designs an extremely efficient network on the benefit of feature reuse. The multi-scale fusion and feature reuse in densely connected architecture make it especially suitable for efficient human pose estimation. But as discussed in~\cite{condense,shufflev2}, original densely connected architecture usually suffers from redundant connections, which will harm the overall efficiency. In this paper, the proposed orthogonal attention block~(OAB) uses a similar architecture, but it is enhanced by mask and channel attention units. These two attention units can explicitly help different layers learn more diverse representations, which largely improves the representing capacity and reduces the proportion of redundant connections. 
	
	\begin{figure}[h]
		\centering
		\includegraphics[width=\linewidth]{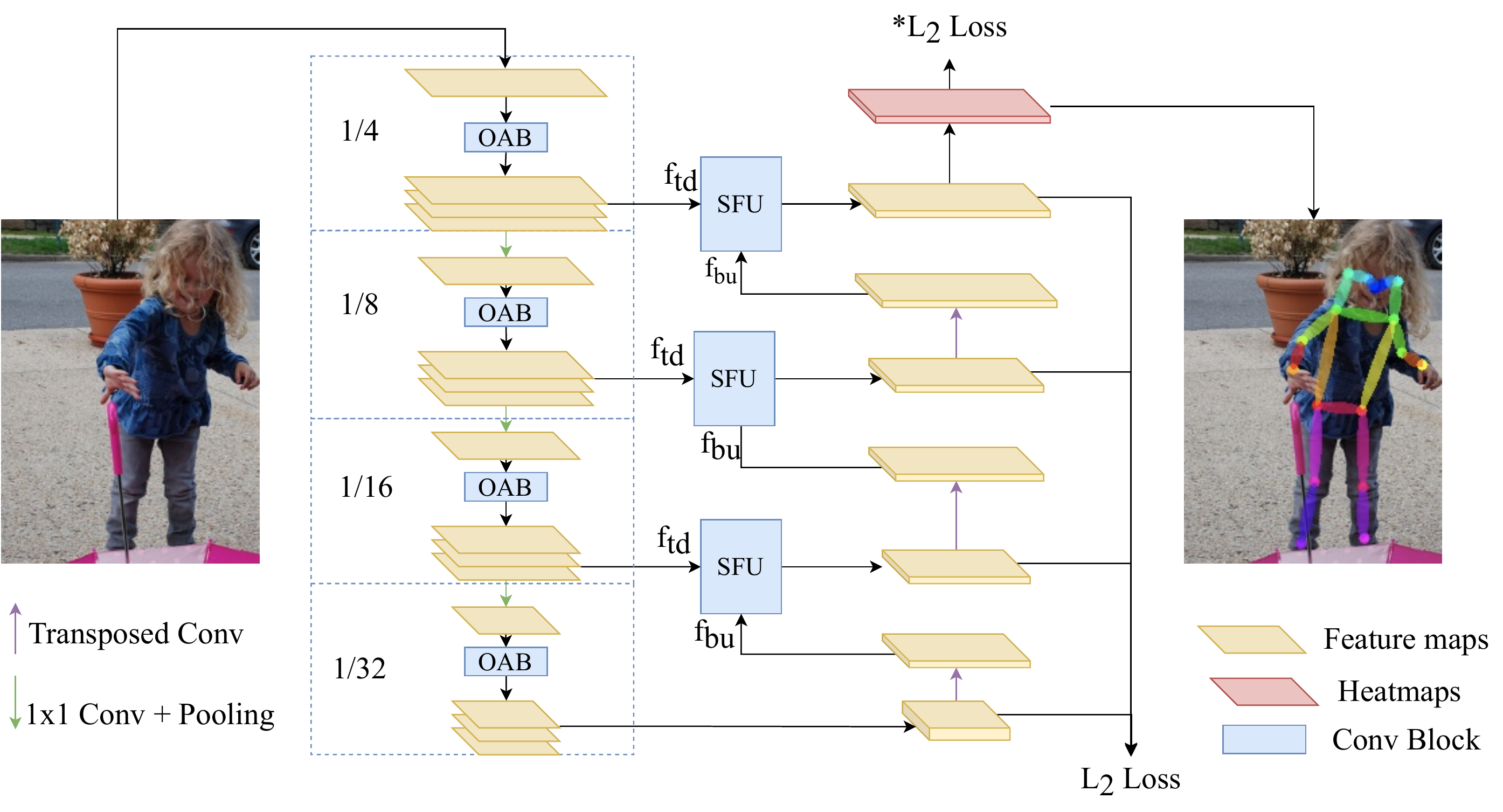}
		\caption{The overview  of DANet. $*$ means online hard keypoints mining (OHKM) \cite{cpn}. \textit{OAB} denotes Orthogonal Attention Block and \textit{SFU} denotes Second-order Attention Unit.}
		\label{DANet}
	\end{figure}
	
	\section{Proposed Method}
	
	In this section, we firstly introduce the overall structure of our network and then go to the details. 
	
	As we have discussed above, we adopt a single pyramid framework. We hope the features of every single spatial size can independently learn rich multi-scale information. We notice that layers within the same spatial size also have different receptive-field sizes. Thus, aggregation among these layers may also be able to exploit multi-scale information. Motivated by this, we propose an orthogonal attention block (OAB). It performs layer-wise aggregation via a densely connected architecture similar to~\cite{densenet}. However, we further enhance it by mask and channel attention units, which can explicitly help different layers to be more diverse and reduce the redundant connections. It will be further discussed in Sec~\ref{OAB_sec}.
	
	As shown in Figure~\ref{DANet}, in the top-down path, OAB extracts a feature pyramid, which contains features with different spatial sizes. And within each spatial size, there are also layers with different receptive-field sizes. These hierarchical features largely enrich multi-scale information. In the bottom-up path, the pyramid features are fused by the proposed second-order fusion unit (SFU), which can further maximize the effective information in the final fused representations. The details will be discussed in Sec~\ref{SFU_Sec}. At last, the deeply aggregated representations are then inputted to a regressor to estimate the heatmaps for all keypoints.
	
	\begin{figure*}[t]
		\centering
		\includegraphics[width=\linewidth]{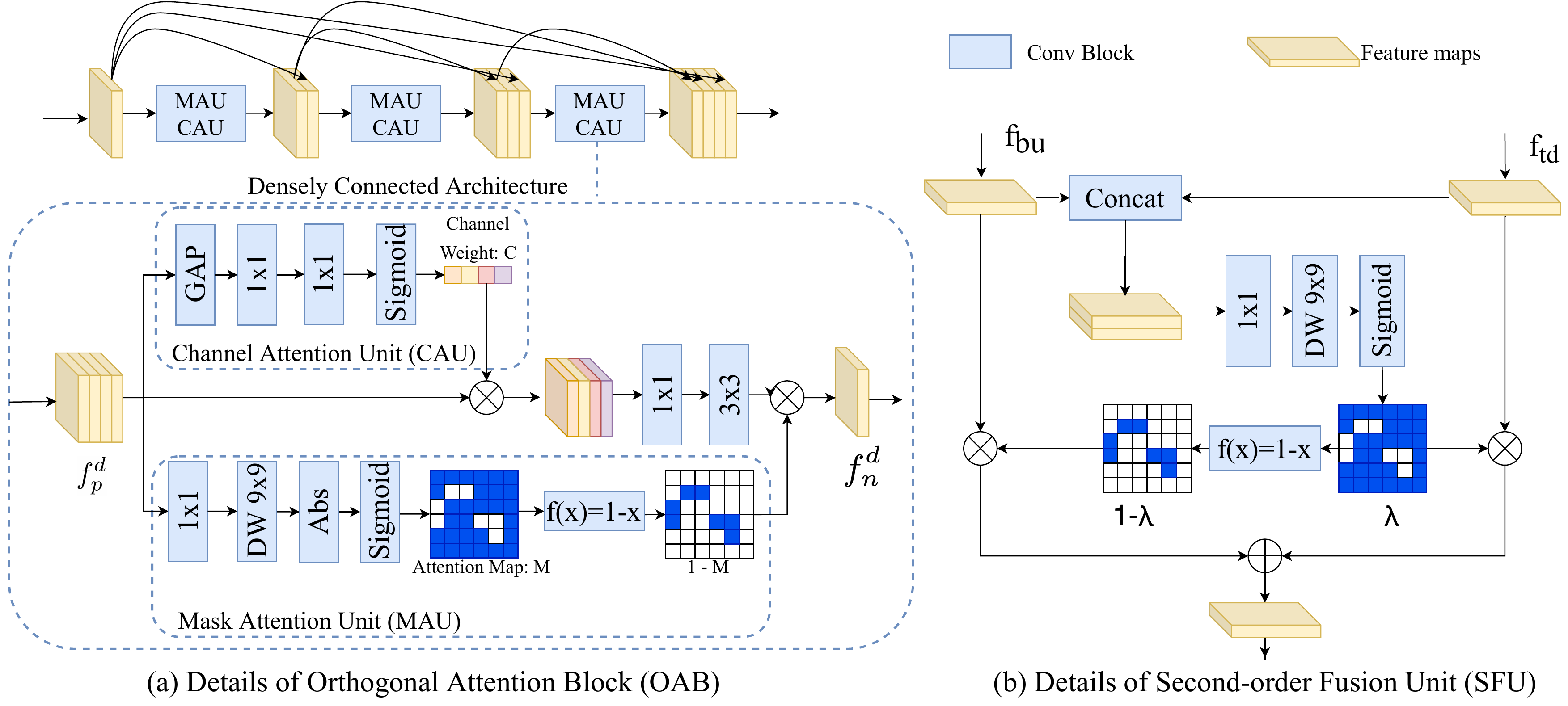}
		\caption{ Details of Orthogonal Attention Block (OAB) and the details of Second-order Fusion Unit (SFU). $k\times k$ means a convolution block composited in the order of $BN-ReLU-conv$. And `DW' means depthwise convolution. `GAP' denotes Global Average Pooling.}
		\label{dense_sfu}
	\end{figure*} 
	
	\subsection{Orthogonal Attention Block}
	\label{OAB_sec}
	As we have discussed in Sec~\ref{densely_connected_arch}, original densely connected architecture suffers from redundant connections. It can perform layer-wise aggregation to exploit multi-scale information from different layers. However, it does not take full advantage of this architecture.
	
	To further explore the benefits of densely connected architecture and alleviate the problem of redundant connections, we hope different layers can learn representations as diverse as possible. Because in this way, the layer-wise aggregated features can contain more effective information and have higher representing capacity. Towards this motivation, two methods are proposed. The first one is explicitly forcing different layers to learn different representations, which is realized via a mask attention unit (MAU). And the second one is improving the nonlinearity of each layer. which is realized via a channel attention unit (CAU).  
	
	\vspace{0.02\linewidth}
	\noindent \textbf{Mask Attention.}
	To explicitly encourage newly generated features to be different from input ones, we propose a mask attention unit (MAU) to mask representations of input features out from output features. Thus, the top features will include less repeated information from bottom layers. As shown in Figure~\ref{dense_sfu}~(a), we firstly extract an attention map , denoted as $M$, which represents representations of input features. Then we multiply $(1-M)$ to the output features to mask out previous representations. To mathematically illustrate this unit, suppose $f^d_p$ denotes input features of the $d^{th}$ layer, and $f^d_n$ denotes the newly generated features. Then in formula,
	\begin{equation}
	\small
	\begin{aligned}
	&M = \psi^d(f^d_p)& &M = Sigmoid(Abs(M)) \\
	&f^d_n = \phi^d(f^d_p)\times(1 - M)& &f^{d+1}_p = Cat(f^d_p, f^d_n),
	\end{aligned}\label{mask_equation}
	\end{equation}
	where $\phi^d(\cdot)$ denotes the mapping function of $d^{th}$ layer, and $\psi^d(\cdot)$ denotes the function used to generate $M$. $Cat(\cdot,\cdot)$ denotes concatenation. Through $Abs(\cdot)$ and $Sigmoid(\cdot)$, $M$ becomes an attention map which highlights pixels with relatively large values. Then $(1-M)$ acts like a mask, which masks representations of $f^d_p$ out from $f^d_n$, and encourages it to be more diverse.
	
	We use a large kernel size, i.e. $9\times9$ in MAU to extract the attention map. Kernels with larger sizes are more robust to local noises and can extract more informative attention. To reduce the computation cost, we use a depthwise convolution to make a trade-off. Similar settings can also be found in \cite{xdense,xunit}. 
	
	\vspace{0.02\linewidth}
	\noindent \textbf{Channel Attention.}
	Motivated by the success of SE unit~\cite{SE} for ResNet \cite{resnet}, we also use a channel attention unit~(CAU) to improve the nonlinearity of each layer. As shown in Figure~\ref{dense_sfu} (a), a channel attention unit is inserted before the $1\times1 $ convolutional layer. By modeling the correlation between different channels, this unit is can effectively improve the overall nonlinearity of original convolutional layers, and help the output feature to be more different from input features.
	
	With both mask and channel attention units, OAB can be mathematically expressed as 
	\begin{equation}
	\small
	\begin{aligned}
	&M = MA(f^d_p) & &C = CA(f^d_p) \\
	&f^d_n = \phi^d(f^d_p\times C) \times(1 - M)& &f^{d+1}_p = Cat(f^d_p, f^d_n),
	\end{aligned}\label{dense_equation}
	\end{equation} 
	where $CA(\cdot)$ denotes the function that generates channel weights $C$, and $MA(\cdot)$ denotes the function that generates attention map $M$. $C$ helps improve the nonlinearity of $\phi^d(\cdot)$, and $1-M$ helps mask representations of $f^d_p$ out from $f^d_n$. With the help of MAU and CAU, OAB can generate layers with diverse representation, which largely enriches the multi-scale information in the single pyramid, and helps DANet generate deeply aggregated representations.
	
	\begin{table*}[t]
		\setlength{\tabcolsep}{0.3cm}
		\centering
		\caption{Details of different DANets.}
		\label{DA_details}
		\resizebox{\linewidth}{!}{
			\begin{tabular}{c|c|c|c|c|c}
				\hline
				Layers &Output Size&DANet-72 & DANet-88 & DANet-98 & DANet-102 \\
				\hline
				Convolution &$128\times96$ &\multicolumn{4}{c}{$7\times7\times64$, stride 2}\\
				\hline
				Convolution &$64\times48$ &\multicolumn{4}{c}{$3\times3\times64$, stride 2} \\
				\hline
				Orthogonal Block (1) &$64\times48$&
				$\left[\begin{aligned} &1\times1\quad32 \\ &3\times3\quad32\end{aligned}\right] \times3$& 
				$\left[\begin{aligned} &1\times1\quad32 \\ &3\times3\quad32\end{aligned}\right] \times4$ & 
				$\left[\begin{aligned} &1\times1\quad32 \\ &3\times3\quad32\end{aligned}\right] \times4$ & 
				$\left[\begin{aligned} &1\times1\quad32 \\ &3\times3\quad32\end{aligned}\right] \times6$ \\
				\hline
				\multirow{2}{*}{Transition Layer (1)} &$64\times48$ & 
				$1\times1\times96$& 
				$1\times1\times128$&
				$1\times1\times128$&
				$1\times1\times128$\\
				\cline{2-6}
				&$32\times24$&  \multicolumn{4}{c}{$3\times3$ MaxPooling, stride 2} \\
				\hline
				Orthogonal Block (2) &$32\times24$&
				$\left[\begin{aligned} &1\times1\quad32 \\ &3\times3\quad32\end{aligned}\right] \times6$& 
				$\left[\begin{aligned} &1\times1\quad32 \\ &3\times3\quad32\end{aligned}\right] \times5$ & 
				$\left[\begin{aligned} &1\times1\quad32 \\ &3\times3\quad32\end{aligned}\right] \times12$ & 
				$\left[\begin{aligned} &1\times1\quad32 \\ &3\times3\quad32\end{aligned}\right] \times12$ \\
				\hline
				\multirow{2}{*}{Transition Layer (2)} &$32\times24$ & 
				$1\times1\times192$& 
				$1\times1\times192$&
				$1\times1\times256$&
				$1\times1\times256$\\
				\cline{2-6}
				&$16\times12$&  \multicolumn{4}{c}{$3\times3$ MaxPooling, stride 2} \\
				\hline
				Orthogonal Block (3) &$16\times12$&
				$\left[\begin{aligned} &1\times1\quad32 \\ &3\times3\quad32\end{aligned}\right] \times12$& 
				$\left[\begin{aligned} &1\times1\quad32 \\ &3\times3\quad32\end{aligned}\right] \times18$ & 
				$\left[\begin{aligned} &1\times1\quad32 \\ &3\times3\quad32\end{aligned}\right] \times16$ & 
				$\left[\begin{aligned} &1\times1\quad32 \\ &3\times3\quad32\end{aligned}\right] \times16$ \\
				\hline
				\multirow{2}{*}{Transition Layer (3)} &$16\times12$ & 
				$1\times1\times384$& 
				$1\times1\times512$&
				$1\times1\times512$&
				$1\times1\times512$\\
				\cline{2-6}
				&$8\times6$&  \multicolumn{4}{c}{$3\times3$ MaxPooling, stride 2} \\
				\hline
				Orthogonal Block (4) &$8\times6$&
				$\left[\begin{aligned} &1\times1\quad32 \\ &3\times3\quad32\end{aligned}\right] \times12$& 
				$\left[\begin{aligned} &1\times1\quad32 \\ &3\times3\quad32\end{aligned}\right] \times14$ & 
				$\left[\begin{aligned} &1\times1\quad32 \\ &3\times3\quad32\end{aligned}\right] \times14$ & 
				$\left[\begin{aligned} &1\times1\quad32 \\ &3\times3\quad32\end{aligned}\right] \times14$ \\
				\hline
				Transition Layer (4)&$8\times6$ & 
				$1\times1\times512$& 
				$1\times1\times640$&
				$1\times1\times640$&
				$1\times1\times640$\\
				\hline
			\end{tabular}
		}
	\end{table*}
	
	\subsection{Second-order Fusion}
	\label{SFU_Sec}
	
	To better utilize the extracted features, we also improve the fusion method. Most existing models fuse the pyramid features by simple summation or a set of learned but still fixed weights. These fusion methods ignore the mutual dependencies between the pyramid features. Thus we propose a second-order fusion unit (SFU) aimed to utilize the second-order relationship. As shown in Figure~\ref{dense_sfu} (b), we firstly concatenate the two input features. Then they are fed to a three-layer convolutional block to generate the weight map $\lambda$, which represents the mutual dependencies between the original two features. These two features are then weighted by $\lambda$ and $1 - \lambda$. Since $\lambda$ and $1-\lambda$ have contrary patterns, the weighted features will be decoupled to more independent ones. In this way, we suppress their similarities while highlighting their diversities. As a result, there will be less redundant information but more effective information in the final fused features.
	
	Suppose $f_{td}$ denotes feature maps from the top-down path and $f_{bu}$ denotes feature maps from the bottom-up path. Then SFU can be expressed as
	\begin{equation}
	\small
	\lambda = Sigmoid(\theta(Cat(f_{td}, f_{bu}))) \quad
	f = \lambda f_{td} + (1 - \lambda) f_{bu},
	\end{equation} 
	where $\theta(\cdot)$ denotes the function that generates aggregating weights. In practice, this function is a combination of pointwise convolution and  depthwise convolution. The kernel size of depthwise convolution is also set as $9$.
	
	\subsection{Network instantiation}
	
	We instantiate DANets by determining the number of layers in OAB, i.e. the number of layers in the top-down path. The \textit{growth rates} for all the densely connected architectures are set as $32$. The output channels of the $1\times1$ bottleneck are also set as $32$. We adjust the output channels and the number of layers for each stage to form models with different model sizes. DANet-$S$ denotes that there are $S$ layers in the top-down path. The details for DANet-72, DANet-88, DANet-98 and DANet-102 are shown in Table~\ref{DA_details}.
	
	\section{Experiments}
	\subsection{Experiments on COCO datasets}
	
	\noindent \textbf{Dataset.} We use COCO train2017 \cite{coco} as our training dataset, which includes 57K images and 150K person instances. We use COCO va2017 (5K images) and COCO test-dev set (20K images) to evaluate our models. The evaluation metric is based on Object Keypoint Similarity (OKS). 
	
	\vspace{0.02\linewidth}
	\noindent\textbf{Training.} We randomly rotate, flip and scale human boxes to augment training instances. The rotation range is $[-45^\circ, +45^\circ]$, and the scaling range is $[0.7,1.35]$. Following \cite{half_body}, half-body transform is also performed for data augmentation. The human boxes will be firstly extended to $4:3$ in terms of hight-width ratio, and then resized to $256\times192$. We use Adam \cite{adam} as our optimizer. The initial learning rate is set as $3\times10^{-4}$ and it decays linearly to $0$ at the end of the training. The batch size is $256$ and we train all models by $2.8\times10^{7}$ iterations no RTX2080TI GPUs. All models are trained on $8$ 2080Ti GPUs.
	
	\vspace{0.02\linewidth}
	\noindent\textbf{Testing.} During testing, we use the same detecting results as~\cite{simplebaseline,hrnet,cpn}, i.e. $56.4$ in score of AP on COCO val2017 set. Following the strategy in \cite{stacked}, images and their flipped versions are all evaluated. Their predicted heatmaps are averaged and then blurred by a Gaussian filter. To obtain the final location, Instead of directly extracting the location of the highest response in heatmaps, we adjust it by a quarter offset in the direction of the second-highest response.
	
	\begin{table*}[t]
		\centering
		\setlength{\tabcolsep}{0.15cm}
		\caption{Comparison of results on COCO validation set. '*' denotes reimplementing results. We cut the depth of HRNet-W32 by half and set its base channel as 16 to form HRNet-W16.}
		\label{val_compare}
		\resizebox{\linewidth}{!}{
			\begin{tabular}{c|c|c|c|c|c|cccccc}
				\hline
				Method &~~Backbone~~&~~Pretrain~~&~~Input Size~~&~~Params (M)~~&~~GFLOPs~~&~~AP~~ &~~AP$^{50}$~~&~~AP$^{75}$~~&~~AP$^{M}$~~&~~AP$^{L}$~~&~~AR~~ \\
				\hline
				8-stage Hourglass \cite{stacked}~~& Hourglass & N & $256\times192$ & $25.2$ & $26.2$ & $66.9$ & - &-&-   & -             & -    \\
				CPN \cite{cpn} & ResNet50 & Y & $256\times192$ & $27.0$ & $6.2$ & $68.6$ & - &-&-   & -             & -    \\
				SimpleBaseline \cite{simplebaseline}~~& ResNet50 & Y & $256\times192$ & $34.0$ & $8.9$ & $70.4$ &$80.6$ &$78.3$&$67.1$&$77.2$&$76.3$\\
				HRNet-W32 \cite{hrnet} & HRNet-W32 & Y & $256\times192$ & $28.5$ & $7.1$ & $\color{red}\mathbf{74.4}$ &$\color{blue}\mathbf{90.5}$ &$\color{red}\mathbf{81.9}$&$\color{red}\mathbf{70.8}$&$\color{red}\mathbf{81.0}$&$\color{red}\mathbf{79.8}$\\
				*MSPN \cite{mspn} &ResNet50& N & $256\times192$ & $95.5$ & $4.7$ & $73.3$ &$90.45$ &$80.3$&$69.2$&$\color{blue}\mathbf{79.4}$&$78.6$\\
				*HRNet-W16& *HRNet-W16 & N & $256\times192$ & $6.9$ & $1.9$ & $71.8$ &$89.5$ &$78.9$&$68.6$&$76.8$&$77.2$\\
				*MSPN \cite{mspn} &ResNet18 & N & $256\times192$ & $11.7$ & $2.5$ & $70.8$ &$89.5$ &$77.3$&$66.6$&$76.9$&$76.2$\\
				*MSPN \cite{mspn}&ResNet34 & N & $256\times192$ & $21.4$ & $4.2$ & $73.1$ &$90.5$ &$80.2$&$69.1$&$79.1$&$78.5$\\
				\hline
				\hline
				DANet-72 & DANet-72 & N & $256\times192$ & $\color{red}\mathbf{3.4}$ &$\color{red}\mathbf{1.0}$ & $71.0$ &$89.6$ &$78.1$&$67.6$&$76.2$&$76.6$\\
				DANet-88 & DANet-88 & N & $256\times192$ & $5.3$ & $1.4$ & $71.9$ &$90.0$ &$79.0$&$68.2$&$77.3$&$77.5$\\
				DANet-98 & DANet-98 & N & $256\times192$ & $5.7$ & $1.6$ & $72.8$ &$90.3$ &$79.9$&$69.2$&$78.3$&$78.4$\\
				DANet-102 & DANet-102 & N & $256\times192$ & $5.8$ & $1.8$ & $\color{blue}\mathbf{73.4}$ &$\color{red}\mathbf{90.6}$ &$\color{blue}\mathbf{80.5}$&$\color{blue}\mathbf{69.9}$&$78.6$&$\color{blue}\mathbf{78.8}$\\
				\hline
			\end{tabular}
		}
	\end{table*}
	\begin{table*}[t]
		\centering
		\setlength{\tabcolsep}{0.15cm}
		\caption{Comparison of results on COCO test-dev set.}
		\label{test_dev_compare}
		\resizebox{\linewidth}{!}{
			\begin{tabular}{c|c|c|c|c|cccccc}
				\hline
				Method~~&~~Backbone~~&~~Input Size~~&~~Params (M)~~&~~GFLOPs~~&~~AP~~&~~AP$^{50}$~~ &~~AP$^{75}$~~&~~AP$^{M}$~~&~~AP$^{L}$~~&~~AR~~ \\
				\hline
				CPN \cite{cpn} &~~ResNet-Inception~~&$384\times288$ &$ - $&$ - $& $72.1$ &$91.4$&$80.0$&$68.7$&$77.2$&$78.5$\\
				SimpleBaseline \cite{simplebaseline}~~& ResNet50  &$256\times192$ & $34.0$ & $8.9$ & $70.0$ &$90.9$ &$77.9$&$66.8$&$75.8$&$75.6$\\
				SimpleBaseline \cite{simplebaseline}~~& ResNet152 &$256\times192$ & $68.6$ & $15.7$ & $71.6$ &$91.2$ &$80.1$&$68.7$&$77.2$&$77.3$\\
				HRNet-W32 \cite{hrnet} & HRNet-W32 & $384\times288$ & $28.5$ & $16.0$ & $\color{red}\mathbf{74.9}$ &$\color{blue}\mathbf{92.5}$ &$\color{red}\mathbf{82.8}$&$\color{red}\mathbf{71.3}$&$\color{red}\mathbf{80.9}$&$\color{red}\mathbf{80.1}$\\
				\hline
				\hline
				DANet-72 & DANet-72 & $256\times192$ & $\color{red}\mathbf{3.4}$ &$\color{red}\mathbf{1.0}$& $70.5$ &$91.8$ &$78.2$&$67.5$&$75.0$&$76.7$\\
				DANet-88 & DANet-88  & $256\times192$ & $5.3$ & $1.4$ & $71.5$ &$91.9$ &$79.5$&$68.5$&$76.0$&$77.6$\\
				DANet-98 & DANet-98 & $256\times192$ & $5.7$ & $1.6$ & $72.6$ &$92.5$ &$80.7$&$69.4$&$77.2$&$78.5$\\
				DANet-102 & DANet-102 & $256\times192$ & $5.8$ & $1.8$ & $\color{blue}\mathbf{72.9}$ &$\color{red}\mathbf{92.6}$ &$\color{blue}\mathbf{80.8}$&$\color{blue}\mathbf{69.8}$&$\color{blue}\mathbf{77.4}$&$\color{blue}\mathbf{78.8}$\\
				\hline
			\end{tabular}
		}
	\end{table*}
	
	\vspace{0.02\linewidth}
	\noindent\textbf{Comparison with Existing Methods.} 
	Intensive experiments are done on DANet-72, DANet-88, DANet-98, and DANet-102. The details of DANets are shown in Table~\ref{DA_details}. Reference models mainly include the families of HRNet~\cite{hrnet}, SimpleBaseline~\cite{simplebaseline} and MSPN~\cite{mspn}. As these models are usually much larger than DANets, we reimplement some of their smaller versions to make a more comprehensive comparison. The results on COCO val2017 and COCO test-dev are shown in Table~\ref{val_compare} and Table~\ref{test_dev_compare} respectively. Our \mbox{DANet-102} has only $1.8G$ FLOPs but achieves comparable results with models over $4G$ FLOPs. DANet-72 has even only $1.0G$ FLOPs, but still achieves $71.0$ AP on COCO val2017 and $70.5$ AP on COCO test-dev. To better illustrate the efficiency of DANets, we plot the AP with respect to the FLOPs of different models. As shown in Figure~\ref{speed}~(a), DANets take up the left corner of the diagram, which means that it can always achieve better accuracy but with smaller model complexity.

	\vspace{0.02\linewidth}
	\noindent\textbf{Ablation Study.}
	We investigate each component, i.e. CAU, MAU, and SFU, in DANet respectively. We remove all components in DANet102 as Baseline1, and then gradually add each component back. Since extra components will introduce additional flops and parameters, to make a more comprehensive comparison, we increase the growth rate in Baseline1 from $32$ to $64$ as Baseline2. The results on COCO val2017 dataset are shown in Table~\ref{ablation_study}. CAU, MAU, and SFU can improve the AP score by $1.5$, $1.9$, and $0.7$ over baseline1 respectively. Although extra flops and parameters are introduced, they have better cost-effectiveness than Baseline2.
	
	\begin{table}[htb]
		\centering
		\setlength{\tabcolsep}{0.1cm}
		\caption{Abation of study of different components in DANet-102. Results are reported on COCO2017 val dataset.}
		\label{ablation_study}
		\resizebox{\linewidth}{!}{
			\begin{tabular}{c|ccc|cc|c}
				\hline
				Components&CAU & MAU & SFU & GFLOPs & Params(M) & AP                          \\
				\hline
				Baseline1 &       &     &     & $1.3$    & $3.4$       & $70.2$                        \\
				Baseline2&        &     &     & $1.8$    & $4.5$       & $71.6$                        \\
				CAU         &\checkmark   &     &     & $1.3$    & $4.8$       & $71.7$                        \\
				MAU         &   & \checkmark   &     & $1.6$    & $4.2$       & $72.1$                        \\
				SFU          &    &     & \checkmark   & $1.5$    & $3.5$       & $70.9$                        \\
				CAU + MAU&\checkmark  & \checkmark  &     & $1.6$    & $5.7$       & $72.6$                        \\
				CAU + MAU  + SFU&\checkmark   & \checkmark   & \checkmark   & $1.8$    & $5.8$       & $73.4$ \\ 
				\hline
		\end{tabular}}
	\end{table} 
	
\begin{figure}[h]
		\centering
		\includegraphics[width=\linewidth]{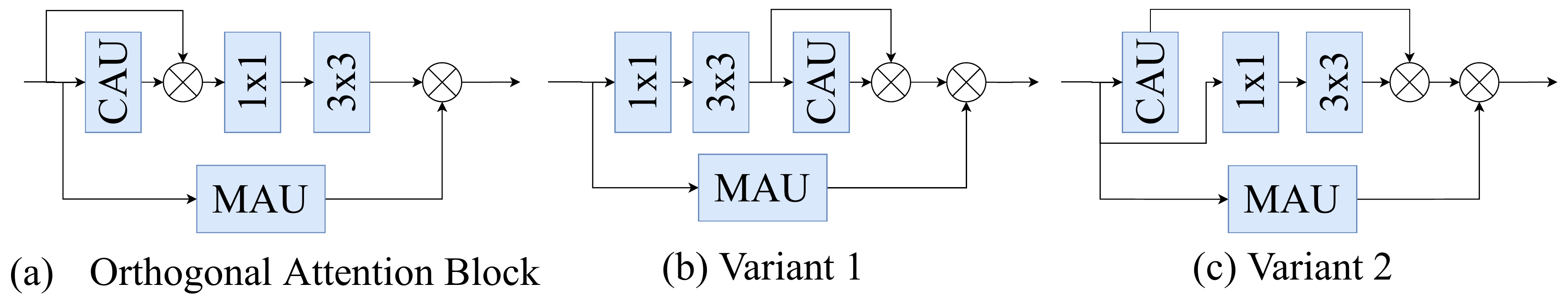}
		\caption{Different variants of channel attention in Orthogonal Attention Blolck.}
		\label{channel_variants}
	\end{figure}

	\vspace{0.02\linewidth}
	\noindent\textbf{Further Discussion about CAU.} 
	The idea channel attention unit (CAU) is firstly proposed in~\cite{SE}. And since its proposal, it has been in used various models. Although they have a similar inner structure, i.e. global pooling followed by two fully connected layers and activation function, the position that CAU is inserted counts a lot. In our case, when MAU is introduced, there are several variants of OAB. As shown in  Figure~\ref{channel_variants}, depending on the position of CAU, OAB has two other variants. Comparative experiments are done between them. Results are shown in Table~\ref{compare_channel_attention}. As one can see, although different variants have the same flops and number of parameters, OAB is the best choice.
	
	\begin{table}[h]
		\centering
		\setlength{\tabcolsep}{0.3cm}
		\caption{Ablation study about variants of OAB. Results are reported on COCO2017 val dataset.}
		\label{compare_channel_attention}
		\resizebox{0.8\linewidth}{!}{
			\begin{tabular}{cccc}
				\hline
				Architecture&GFLOPs&Params&AP\\
				\hline
				Variant 1 &$1.6$&$5.7$&$72.3$\\
				Variant 2 &$1.6$&$5.7$&$72.3$\\
				OAB &$1.6$&$5.7$&$\mathbf{72.6}$ \\
				\hline
		\end{tabular}}
	\end{table}
	
	\vspace{0.02\linewidth}
	\noindent\textbf{Further Discussion about MAU.} 
	Based on densely connected architecture, we further prove the superiority of mask attention unit~(MAU). Generally speaking, mask attention unit can be regarded as a kind of spatial attention unit, but it is more elaborate designed. To validate the details in MAU, we compare it with other kinds of variants. We leave out the CAU and SFU in DANet-102 as the Baseline1. Since attention units will introduce extra complexity, we modify the growth rate of densely connected architecture from $32$ to $64$ to form Baseline2. The first variant is called xunit~\cite{xunit}, which is also used in DxNet \cite{xdense}. We  also modify $f^d_n = \phi^d(f_p)\times(1 - M)$ in Equation~1 to $f^d_n = \phi^d(f_p)\times M$ and $f^d_n = \phi^d(f_p)\times(1 + M)$ to form MAU-Variant1 and MAU-Variant2 respectively. 
	
	As shown in Table~\ref{compare_spatial_attention},  despite that MAU introduces extra model complexity over Baseline1, it outperforms Baseline2 by a large margin with smaller model size and complexity. And among all kinds of spatial attention units, MAU also has the best performance. Variant1 and Variant2 have the same complexity as MAU but have inferior performances. This comparison indicates that every detail in MAU plays an important role.
	
	\begin{table}[h]
		\centering
		\setlength{\tabcolsep}{0.2cm}
		\caption{Ablation study of different spatial attentions. Results on COCO validation set are reported.}
		\label{compare_spatial_attention}
		\resizebox{0.9\linewidth}{!}{
			\begin{tabular}{cccc}
				\hline
				Methods&GFLOPs&Params (M)&AP\\
				\hline
				Baseline1 &$1.3$&$3.4$&$ 70.2$ \\
				Baseline2 &$1.8$&$4.5$&$ 71.6$\\
				xunit \cite{xdense} &$1.4$&$3.6$&$ 71.6$ \\
				MAU-Variant1 &$1.6$&$4.2$&$71.8$\\
				MAU-Variant2 &$1.6$&$4.2$&$71.9$\\
				MAU  &$1.6$&$4.2$&$\mathbf{72.1}$\\
				\hline
			\end{tabular}
		}
	\end{table}
	
	\vspace{0.02\linewidth}
	\noindent\textbf{Further Discussion about SFU.} 
	The proposed second-order fusion unit can maximize the effective information in the final fused features. To prove that it works with different kinds of input features, ablation studies are done on different backbones. Three backbone are chosen, i.e. Res18~\cite{resnet}, DANet-102 without MAU and CAU, and DANet-102. We investigate their performances when the pyramid features are fused by summation or SFU. As shown in Table~\ref{ablation_convex}, with different backbones, SFU can always improve the performance by a large margin. Especially, with the help of SFU, DANet-102 can learn further deeply aggregated representations, and the performance is improved by $0.8$ in the score of AP, which makes the extra $0.2G$ FLOPs worthwhile.
	
	\begin{table}[h]
		\label{ablation_convex}
		\setlength{\tabcolsep}{0.2cm}
		\caption{Ablation study of second-order fusion. Results on COCO validation set are reported.}
		\resizebox{\linewidth}{!}{
			\begin{tabular}{c|c|ccc}
				\hline
				Backbone&Fusion&GFLOPs&Params (M)&AP\\
				\hline
				\multirow{2}{*}{Res18}&Sum&$2.5$&$11.8$& $70.8$ \\ 
				&SFU &$2.7$&$11.9$& $\mathbf{71.4}$ \\ 
				\hline
				\multirow{2}{*}{\makecell[c]{DANet-102 \\ (No MAU and CAU)}}&Sum&$1.3$&$3.4$&$ 70.2$ \\
				&SFU&$1.5$&$3.5$&$\mathbf{70.9}$ \\
				\hline
				\multirow{2}{*}{DANet-102}&Sum&$1.6$&$5.7$&$72.6$\\
				&SFU&$1.8$&$5.8$&$\mathbf{73.4}$\\
				\hline
			\end{tabular}
		}
	\end{table}
	
	\begin{table*}[t]
		\centering
		\setlength{\tabcolsep}{0.3cm}
		\caption{Results on MPII test dataset. }
		\label{mpii}
		\resizebox{\linewidth}{!}{
			\begin{tabular}{c|c|c|ccccccc|c}
				\hline
				Methods &GFLOPs&Params(M)&Head&Shoulder&Elbow&Wrist&Hip&Knee&Ankle&Mean\\
				\hline
				Newell et al. \cite{stacked}&$55$&$26$&$98.2$&$96.3$&$91.2$&$87.1$&$90.1$&$87.4$&$83.6$&$90.9$\\
				Ning et al. \cite{coco}&$124$&$74$&$98.1$&$96.3$&$92.2$&$87.8$&$90.6$&$87.6$&$82.7$&$91.2$\\
				Peng et al. \cite{peng2018jointly}&$55$&$26$&$98.1$&$96.6$&$92.5$&$\color{blue}\mathbf{88.4}$&$90.7$&$87.7$&$83.5$&$91.5$\\
				Chu et al. \cite{chu2017multi}&$128$&$58$&$98.5$&$96.3$&$91.9$&$88.1$&$90.6$&$88.0$&$\color{blue}\mathbf{85.0}$&$91.5$\\
				Yang et al. \cite{yang2017learning}&$46$&$28$&$\color{red}\mathbf{98.5}$&$\color{red}\mathbf{96.7}$&$\color{red}\mathbf{92.5}$&$\color{red}\mathbf{88.7}$&$\color{red}\mathbf{91.7}$&$\color{red}\mathbf{88.6}$&$\color{red}\mathbf{86.0}$&$\color{red}\mathbf{92.0}$\\
				Sekii et al. \cite{pose_proposal}&$6$&$16$&$-$&$-$&$-$&$-$&$-$&$-$&$-$&$88.1$\\		
				FPD \cite{fast_human}&$9$&$\color{red}\mathbf{3}$&$98.3$&$96.4$&$91.5$&$87.4$&$90.9$&$87.1$&$83.7$&$91.1$\\
				\hline
				\hline
				DANet-72  &$\color{red}\mathbf{1.6}$&$3.4$&$97.9$&$96.3$&$90.9$&$87.0$&$90.7$&$87.1$&$82.3$&$90.6$\\		
				DANet-88  &$2.8$&$5.3$&$98.0$&$96.4$&$91.5$&$87.3$&$90.8$&$87.6$&$82.9$&$91.0$\\		
				DANet-98  &$2.2$&$5.7$&$98.1$&$96.5$&$92.1$&$87.6$&$90.8$&$88.2$&$83.5$&$91.4$ \\
				DANet-102&$2.4$&$5.8$&$98.1$&$\color{red}\mathbf{96.7}$&$\color{blue}\mathbf{92.3}$&$87.9$&$\color{blue}\mathbf{91.1}$&$\color{blue}\mathbf{88.4}$&$84.0$&$\color{blue}\mathbf{91.6}$\\
				\hline
		\end{tabular}}
	\end{table*}
	
	\vspace{0.02\linewidth}
	\noindent\textbf{Densely Connected Structure.} 
	In fact, the densely connected structure also plays an important role in our network. Its layer-wise aggregation scheme enables the network to exploit multi-scale information from layers with different receptive-field sizes. In this section, we experimentally prove that densely connected structure is more suitable for HPE than commonly used ResNet~\cite{resnet}.  A classical single pyramid network without any extra technical skills is selected as our baseline. We replace the ResNet~\cite{resnet} backbone with DenseNet~\cite{densenet}. As shown in Table~\ref{ablation_dense}, DenseNet can always achieve better results than ResNet, which indicates that the multi-scale information from layers with different receptive-field sizes may help improve the performance of a single pyramid network.
	
	\begin{table}[h]
		\centering
		\setlength{\tabcolsep}{0.3cm}
		\caption{The results of single pyramid framework with different backbones on COCO2017 val dataset.}
		\label{ablation_dense}
		\resizebox{0.8\linewidth}{!}{
			\begin{tabular}{cccc}
				\hline
				Backbone&GFLOPs&Params (M)&AP\\
				\hline
				Res18&$2.5$&$47.0$& $70.8$ \\ 
				Dense98&$2.4$&$26.3$& $\mathbf{72.7}$ \\ 
				\hline
				Res34&$4.2$&$85.5$&$73.1$ \\
				Dense121 &$3.0$&$33.4$&$\mathbf{73.3}$ \\
				\hline
				Res50&$4.7$&$95.4$&$73.3$\\
				Dense169&$3.6$&$57.4$&$\mathbf{73.6}$\\
				\hline
			\end{tabular}
		}
	\end{table}
	
\begin{figure}[h]
		\centering
		\includegraphics[width=\linewidth]{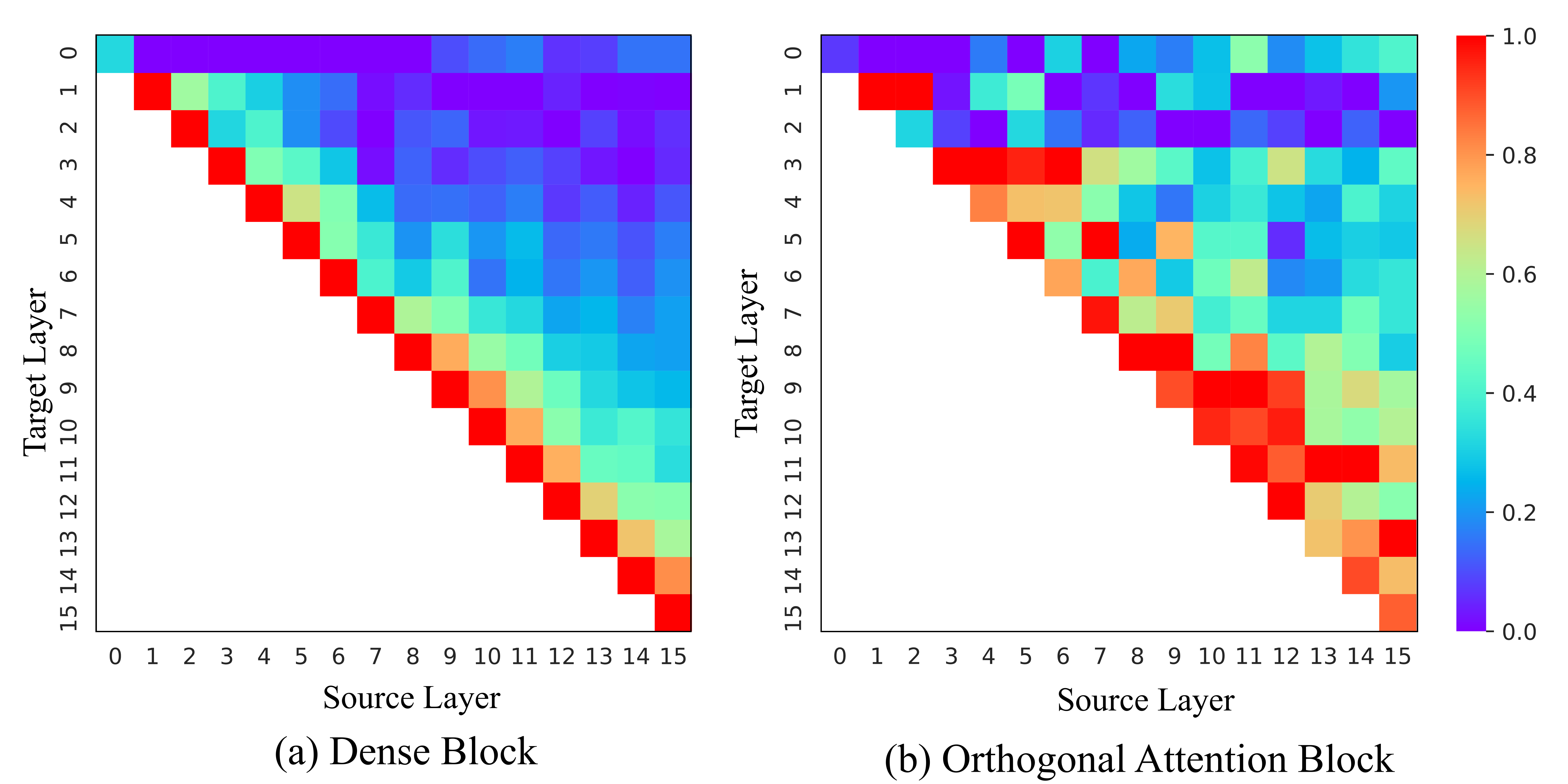}
		\caption{Illustration of weight distribution among different layers. The color of pixel ($s$, $l$) encodes the average L1-norm of weights connecting layer $s$ to $l$. (a) All layers in Dense block distribute most weights to only latest layers and result in redundant connections. (b) With mask and channel attention units, the redundant connections are largely reduced.}
		\label{weight_distribution}
	\end{figure}

	\vspace{0.02\linewidth}
	\noindent\textbf{Diverse Representations.} 
	We analyze the diversities and redundancies by visualizing convolutional weights that connect different layers in Figure~\ref{weight_distribution}. The idea is intuitive. Less diverse features tend to get smaller convolutional weights, i.e. more redundant connections because connecting such features take no information gain. As shown in Figure~\ref{weight_distribution}, most connections in the original dense block have only negligible weights, while more connections in OAB have weights of significant values. It indicates that OAB may produce more diverse features.
	
	\subsection{Inference Speed}
	
	To directly illustrate the efficiency of DANet, the inference speed of different models are evaluated on the same CPU platform with $8$ CPUs (Intel(R) Xeon(R) Gold6013@2.1GHZ). The inference speed is indicated by the number of persons that the model can process in one second, i.e. Persons-Per-Second (PPS). The  input sizes are $256\times192$. For each model, we report the average speed of processing $1\times10^3$ persons. We systematically examined nowadays most popular HPE networks, including the family of HRNet \cite{hrnet} and SimpleBaseline \cite{simplebaseline}, and 8-stage Hourglass \cite{stacked}. The inference speed with respect to the accuracy on COCO validation set is shown in Figure~\ref{speed} (b). Our DANet-102 achieves 73.4 AP with the speed being about $32$ PPS.  DANet-72 has a fast speed being about $58$ PPS, and its AP score is $71.0$. Compared with other models, our DANets make a better trade-off between accuracy and inference speed.
	
	\subsection{Experiments on MPII}
	
	We also train and evaluate our models on MPII \cite{mpii}. It contains 25K scene images and 40K annotated persons. We use the standard train/val/test data split (26K for training, 3K for validation, and 11K for testing) \cite{tompson2014joint}. The results are evaluated by Percentage of Corrected Keypoints (PCK), which indicates the fraction of correct predictions within an error threshold $\tau$ ($\tau = 0.5$ for MPII, i.e. PCKh@0.5). The results are shown in Table~\ref{mpii}. The computation cost of DANets is far smaller than that of other models, but the performances of DANets are comparable with models over 40G FLOPs.
	
	\subsection{Experiments on ImageNet}
	\label{imagenet_sec}
	To further prove the effectiveness of orthogonal attention block (OAB), we also compare it with other densely connected architectures on  ILSVRC 2012 classification dataset~\cite{imagenet}. We replace the dense block in DenseNet169 by OAB, denoted as OAB-DenseNet169, and compare it with DenseNet201 and DxNet169 \cite{xdense}. Their results on test dataset are shown in Table~\ref{imagenet}. The accuracy of OAB-DenseNet169 is 1.3 higher than that of DenseNet169 and even slightly higher than DenseNet201. These results suggest that OAB can largely boost the performance of densely connected architecture, not only for human pose estimation but also potentially for other tasks.
	\begin{table}[h]
		\centering
		\setlength{\tabcolsep}{0.15cm}
		\caption{Results of on ImageNet2012 test dataset. Top-1 and top-5 accuracies with single-crop testing are reported.}
		\label{imagenet}
		\resizebox{\linewidth}{!}{
			\begin{tabular}{ccccc}
				\hline
				Model &GFLOPs&Params (M)&Top-1(\%)&Top-5(\%)\\
				\hline
				DenseNet169 \cite{densenet} &$3.1$&$13.5$&$23.80$&$6.85$\\
				DenseNet201 \cite{densenet} &$4.0$&$19.1$&$22.58$&$6.34$\\
				DxNet169 \cite{xdense}&$3.3$&$13.8$&$23.90$&$7.01$\\
				OAB-DenseNet169&$3.6$&$18.0$&$\mathbf{22.50}$&$\mathbf{6.32}$\\
				\hline
		\end{tabular}}
	\end{table}
	
	\section{Conclusion}
	
	Most existing HPE models use a cascaded framework to extract rich multi-scale information. This framework makes these networks very deep and complex. In this paper, we focus on exploiting multi-scale information from layers with different receptive-field sizes and then making full use of this information by improving the fusion method. The proposed OAB learns multi-scale information from different layers and enhances them by encouraging them to be diverse. The SFU adaptively selects and fuses diverse multi-scale features to maximize the effective information in final fused representations. With the help of OAB and SFU, our single pyramid network can generate deeply aggregated representations that contain even richer multi-scale information and have a larger representing capacity than that of cascaded networks. Specifically, \mbox{DANet-72} achieves $70.5$ AP on COCO test-dev set, with $1.0G$ FLOPs. The inference speed on a CPU platform can reach about $58$ PPS.
	
	\bibliographystyle{IEEEtran}
	\bibliography{pose}
	
\end{document}